\documentclass[conference]{IEEEtran}
\IEEEoverridecommandlockouts
\usepackage{cite}
\usepackage{amsmath,amssymb,amsfonts}
\usepackage{algorithmic}
\usepackage{graphicx}
\usepackage{textcomp}
\usepackage{xcolor}
\usepackage{subcaption}
\usepackage[hyphens]{url}
\usepackage{hyperref}
\hypersetup{breaklinks=true}
\usepackage{amsmath}

\def\BibTeX{{\rm B\kern-.05em{\sc i\kern-.025em b}\kern-.08em
    T\kern-.1667em\lower.7ex\hbox{E}\kern-.125emX}}
\begin{document}

\title{ELIMINATION\\from Design to Analysis}

\author{
\IEEEauthorblockN{Ahmed Khalifa, Dan Gopstein, Julian Togelius}
\IEEEauthorblockA{
\textit{Game Innovation Lab, New York University}, NY, USA\\
ahmed@akhalifa.com, dgopstein@nyu.edu, julian@togelius.com
}
}

% \author{
% \IEEEauthorblockN{Ahmed Khalifa}
% \IEEEauthorblockA{\textit{Tandon School of Engineering} \\
% \textit{New York University}\\
% New York, USA \\
% ahmed@akhalifa.com}
% \and
% \IEEEauthorblockN{Dan Gopstein}
% \IEEEauthorblockA{\textit{Tandon School of Engineering} \\
% \textit{New York University}\\
% New York, USA \\
% dgopstein@nyu.edu}
% \and
% \IEEEauthorblockN{Julian Togelius}
% \IEEEauthorblockA{\textit{Tandon School of Engineering} \\
% \textit{New York University}\\
% New York, USA \\
% julian@togelius.com}
% }

\maketitle

\begin{abstract}
Elimination is a word puzzle game for browsers and mobile devices, where all levels are generated by a constrained evolutionary algorithm with no human intervention. This paper describes the design of the game and its level generation methods, and an analysis of playtraces from almost a thousand users who played the game since its release. The analysis corroborates that the level generator creates a sawtooth-shaped difficulty curve, as intended. The analysis also offers insights into player behavior in this game.
\end{abstract}


\section{Introduction}
Using procedural content generation (PCG) in games, especially for core game elements such as levels, is a bit of an art. When it is executed well, it enhances the player experience and keeps the player wanting to play again~\cite{glagowski2018isaac} but any mistake in the system and the game could feel boring and repetitive, and might lead to disappointment~\cite{maiberg2016nomanssky}. Just like game design, PCG systems are designed using designers' intuition and a trial and error process. Designers usually tests the system on a group of players and based on the feedback they adjust the generator until it works as intended~\cite{yu2016spelunky}. 

The designers of successful games often write a postmortem about what went right and what went wrong during the development process to summarize the lessons learned and help themselves and others to better understand the underlying systems~\cite{yu2016spelunky,mcmilleni2012isaac,lee2014rogue}. Few postmortems, however, are supplemented by player data to validate the design decisions and understand players; conversely, game analytics paper are rarely written by the designers of the analyzed game~\cite{weber2011modeling,drachen2013game}. In this paper, we will discuss the design decisions for the word puzzle game \emph{Elimination} and its level generator. Our postmortem is complemented by an analysis of player data to validate our generation system and better understand the parameters influencing player decisions.

% Procedural content generation (PCG) refers to a computer program that is being used to generate a certain type of content. This content can varies between levels, music, etc. PCG algorithms are typically developed to work for a specific game but recently we have been seeing a new trend of developing PCG algorithms for a known genre~\cite{khalifa2015automatic} or may for any game~\cite{khalifa2016general}. Procedural level generation has been used in video game industry since early days to provide new and unique RPG experiences such as in Beneath Apple Manor (Don Worth, 1978) and Rogue (Glenn Wichman, 1980), or to provide huge amount of content on a small footprint such as in Elite (David Braben and Ian Bell, 1984). 

\section{The Design of Elimination}
Elimination is a word puzzle game that was designed during GameZanga 8\footnote{\url{https://itch.io/jam/gamezanga8}}, a game jam, with ``Minimalist'' as the theme of the jam. The game follows the theme literally by showing the player a bunch of letters and the player has to remove some of these letters so that the remaining letters form a recognized English word. In this paper, we are analyzing the final version of the game which was released on January 23, 2019 for browsers, iOS, and Android devices. The game supports three different gameplay modes: Infinite, Daily, and Levels. In this work, we will only focus on the Levels mode, which presents players with a fixed sequence of levels, as more people played it. It also has a fixed goal (beating the levels), unlike the other two modes. The game includes 30 levels with 10 challenges each, all of which are procedurally generated.
% each which will be described in details in the following subsections.

\subsection{Challenges}
\begin{figure*}
    \centering
    \begin{subfigure}{0.3\linewidth}
        \centering
        \includegraphics[width=\linewidth]{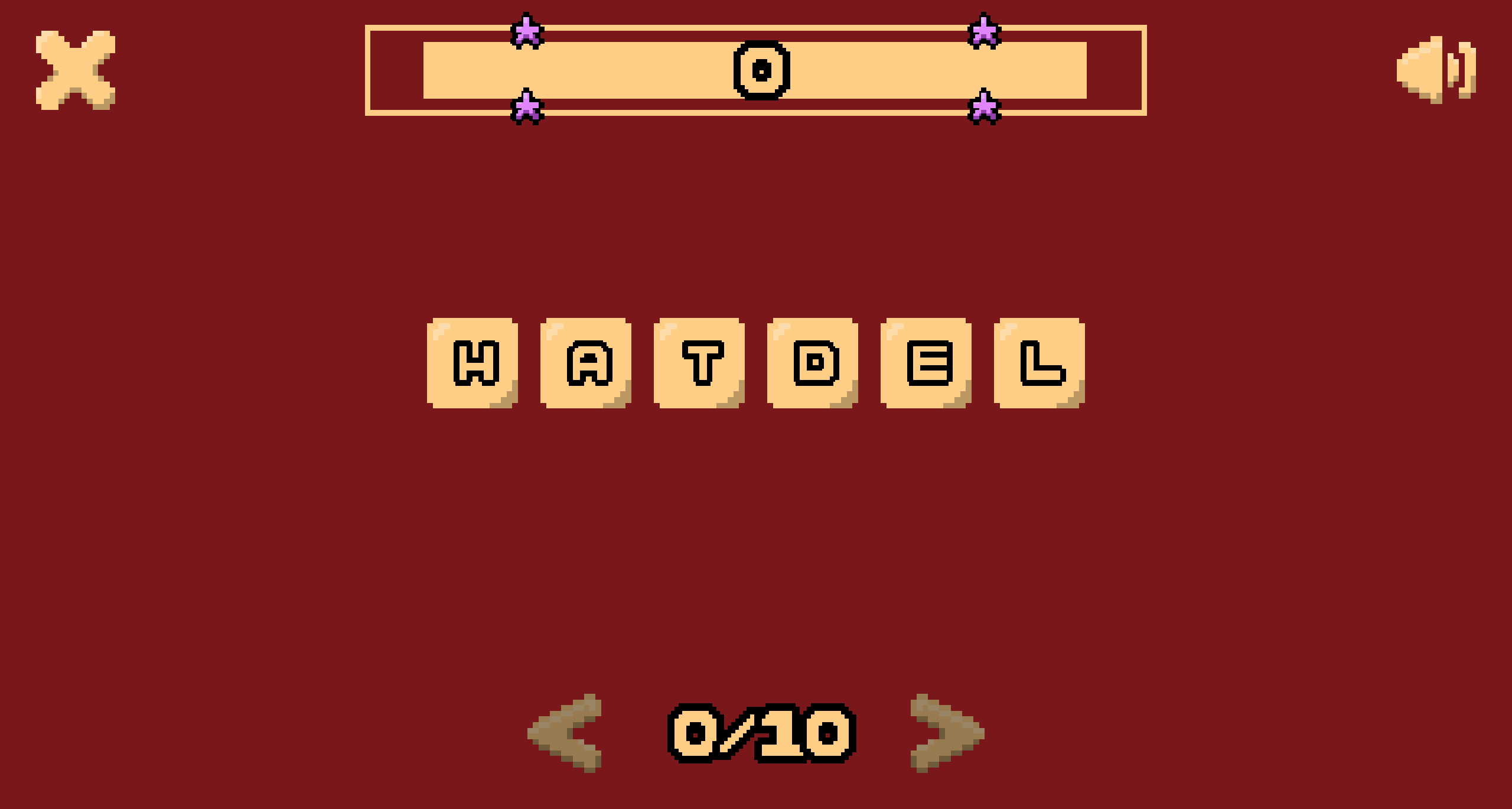}
        \caption{Current Challenge.}
        \label{fig:challenge_a}
    \end{subfigure}
    \begin{subfigure}{0.3\linewidth}
        \centering
        \includegraphics[width=\linewidth]{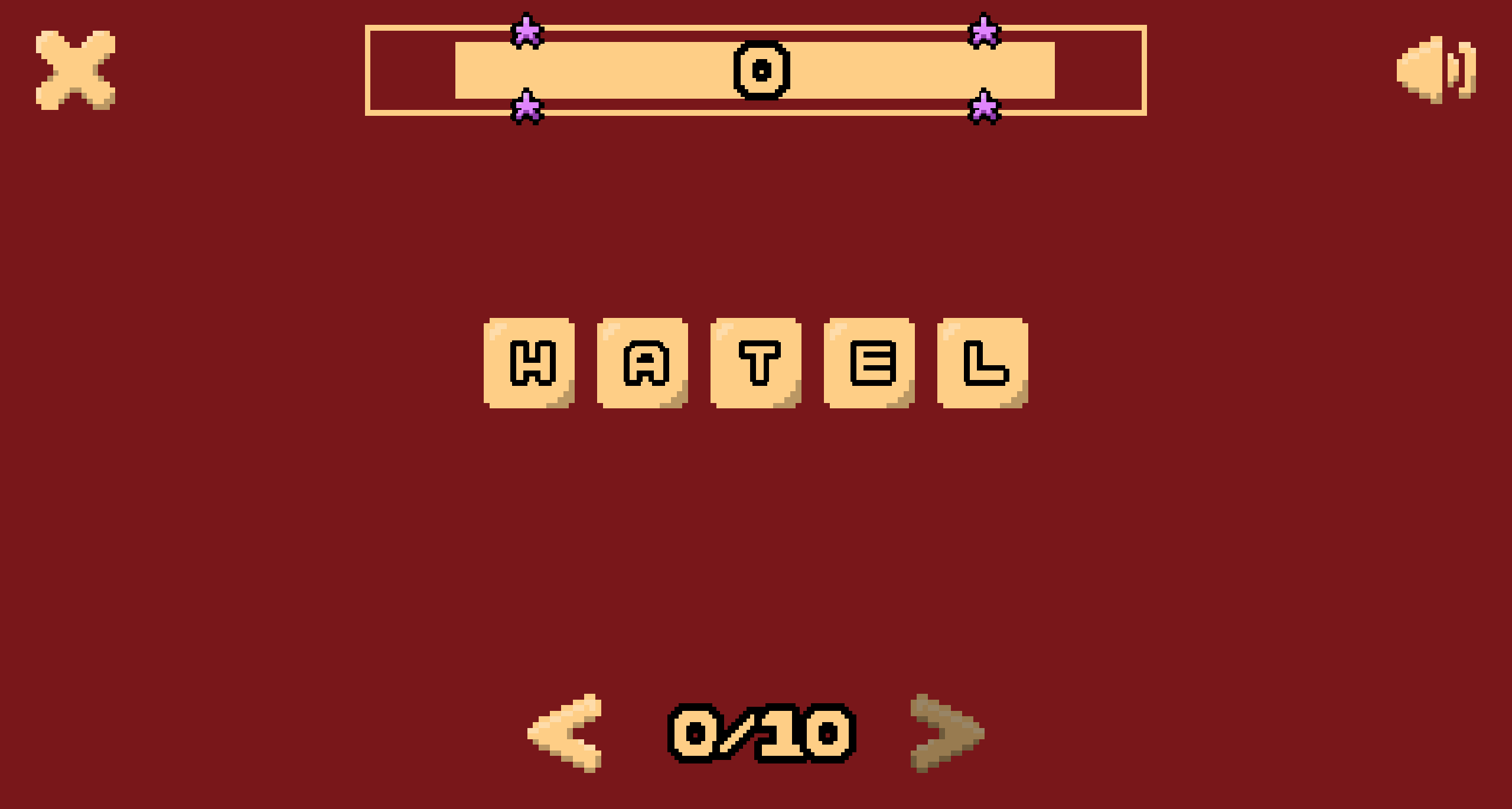}
        \caption{Eliminate ``D'' letter.}
        \label{fig:challenge_b}
    \end{subfigure}
    \begin{subfigure}{0.3\linewidth}
        \centering
        \includegraphics[width=\linewidth]{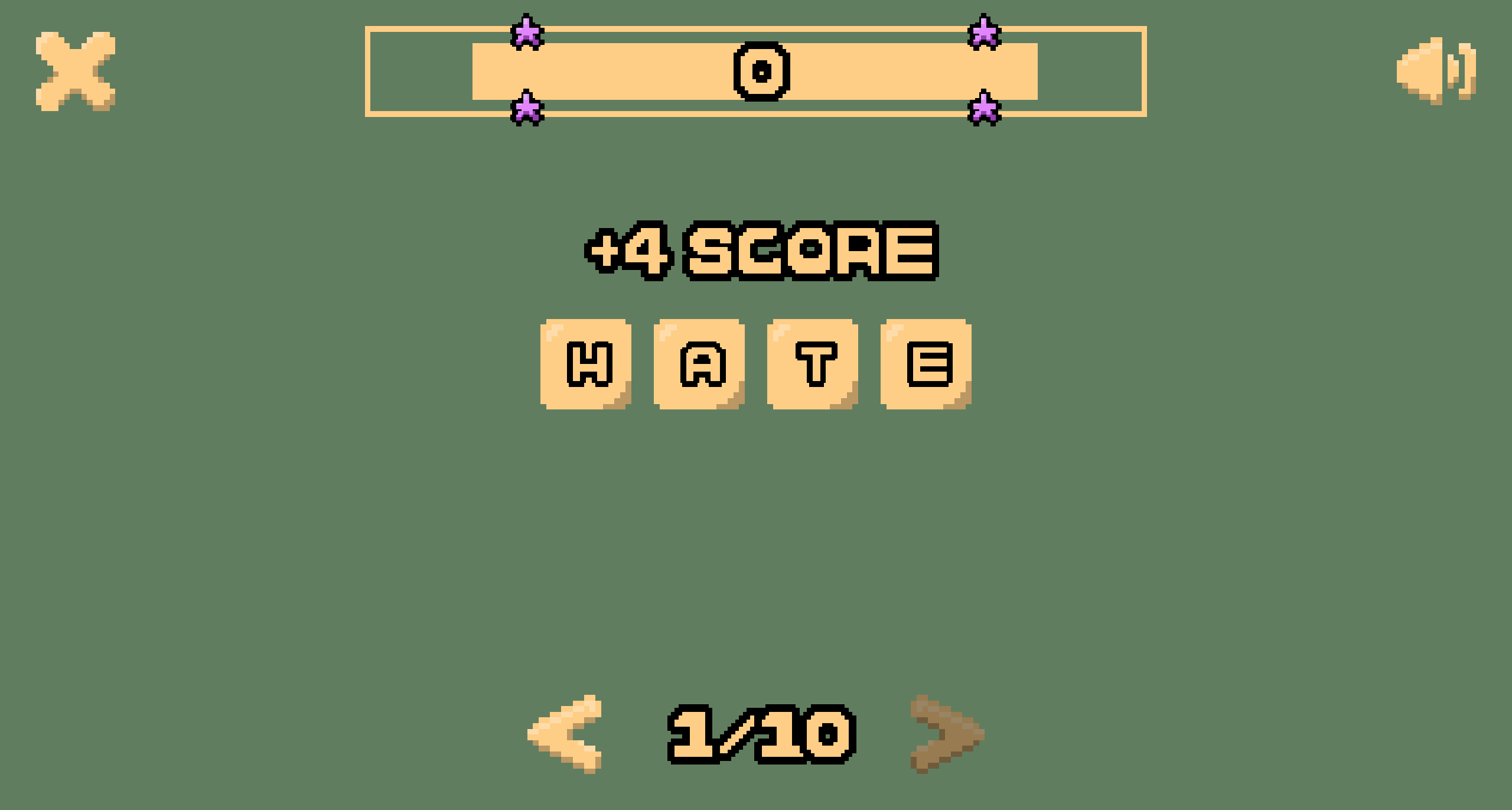}
        \caption{Eliminate ``L'' letter.}
        \label{fig:challenge_c}
    \end{subfigure}
    \caption{Example of a challenge where the player eliminated ``D'' and ``L'' letters to find the word ``HATE'' for 4 points of score.}
    \label{fig:challenge}
\end{figure*}

An Elimination challenge is a string of letters which doesn't form a word by itself but where removing some of the letters will result in a recognized English word. A challenge should have more than one word that can be discovered after removing a group of letters. Figure~\ref{fig:challenge} shows an example of a challenge in the game. The player is faced with 6 letters ``HATDEL'' (shown in figure~\ref{fig:challenge_a}). The player can eliminate letters by clicking them. A solution for that challenge can be seen in figure~\ref{fig:challenge_b} and \ref{fig:challenge_c} respectively where the player removes the letter  ``D'' then ``L'' to find ``HATE''. That is not the only solution as the player can remove ``H'', ``D'', and ``L'' to find ``ATE'' 
%or ``D'', ``E'', and ``L'' to find ``HAT''
. One of the incentives to lead the player to find a certain word than others is that the score depends on the length of the found word. Looking back on the example in figure~\ref{fig:challenge}, ``HATE'' will have score of 4 while ``ATE'' 
%and ``HAT'' 
will have score of 3 on the other hand.

% \begin{figure}
%     \centering
%     \includegraphics[width=0.53\linewidth]{images/Score2X.png}
%     \caption{The score is equal to the length of word multiplied by 2 if one of the letters is carrying 2X symbol.}
%     \label{fig:challenge2X}
% \end{figure}

We also use bonus letters which are called the 2X letter to influence the player choices.
% Another game design trick, we decided to use to influence the player choice for some words is using bonus letters which we call the 2X letter. 
The 2X letter multiplies the score of the found word by 2. For example:
% figure~\ref{fig:challenge2X} shows that 
if the player found the word ``TRUE'' where the ``U'' is a 2X letter the resulted score will be 8 instead of 4. The 2X letter is a good way to influence the player choices toward certain words which could be used in creating easier levels for the player.

\subsection{Levels}
An Elimination level consists of 10 consecutive challenges where each challenge has to be solved before a timer runs out. A level ends either when the player find the 10th word or if the timer runs out. The timer is added to make sure that the players don't solve the challenges by using brute force and listing all the different possible words and then pick the highest score. The timer also adds an element of pressure to a generally casual and relaxed game. Equation~\ref{eq:challengeTime} shows the time in seconds allocated for each challenge in the level where $challengeNumber$ is the index of the challenge which range between 1 and 10. 
% This equation makes sure the first challenge is around 30 seconds while the 10th challenge have around 10 seconds.

\begin{equation}\label{eq:challengeTime}
    challengeTime = \frac{30}{1 + (challengeNumber - 1)/5}
\end{equation}

When the level ends (regardless of whether all 10 words are found), the game automatically unlocks the next level for the player to be played. We decided to allow progress to make sure the game is more relaxing and casual by not getting players stuck on particular level. This will make sure that players who play the same level are doing so willingly and not because the game forces them to.

% \subsection{Metrics}
% There are three different success metrics for any level in Elimination: number of words, score, and time. Number of words is equal to the number of found words in the level which ranges between 0 and 10 where 0 means the timer ran out on the first challenge and 10 means the player beaten all 10 challenges. Score is the score the player got from playing the level. As described before the score depends on the length of letters in the each found word and if there is a 2X letter. Time is number of words that the user found before 25\% of the timer ran out. This makes time also a value between 0 and 10 where 0 means that the player didn't find any word quickly enough and 10 means that the player found all words in very small amount of time (25\% of the timer).

% The player gets a star for every metric that they are able to get the maximum value in it to reward different play styles by players. Number of words rewards players that want to finish the level regardless of playing efficient or fast. Score rewards players that pick words with the highest score. Time rewards players that can spot words fast enough in the game.

\section{Puzzle generation}
Elimination levels are generated by running an evolutionary algorithm for 10 times with same parameters to generate a a single challenge per run. No human curation or editing was performed when generating the included levels. 
% Elimination levels are all generated either on the fly for the Infinite Mode or offline for the Levels and the Daily Mode. In this section, we will focus on the offline generator as it is used in the Levels mode but the online generator is exactly the same but with more loose optimization.

\subsection{Generative System}
For the generation we used Feasible Infeasible 2-Population (FI2Pop) Algorithm~\cite{kimbrough2008feasible} to generate the game challenges. Each challenge is generated by mixing a group of English words called the source words (usually two or three words) together to form the challenge word. The challenge here is that the mixture has to be efficient and this is tested by satisfying two basic constraints: the final word has to be less than a target length and the order of the letters in the challenge has to obey the same order in the source words. The first constraint is added to make sure the system do an intelligent mixing of the source words so there is no repeated letters. For example: ``CAR'' and ``COOL'' could be mixed as ``CARCOOL'' or it can be mixed as ``CAROOL'' where the second one is a better mixing than the first as it has fewer repeated letters. The second constraint is added to make sure that the player can solve the level and find any of the source words. For example: if ``CDATOG'' is a mixture of the source words ``CAT'' and ``DOG'', the `G' can't come before the `O' as it will be impossible to construct ``DOG'' from that mixture by just eliminating letters. 

Based on that, we decided that the chromosome consists of a list of integer values. The first three values point to the source words in the corpus, while the remaining values are used for the mixing procedure. The mixing procedure uses the integer values to determine which mixing word to grab the next letter from. After the mixing is done (before testing for constraints or fitness), the challenge word is shortened by using a greedy algorithm that removes a single letter at a time, each time verifying that the source words can be still extracted from the new challenge word. The reduction algorithm continues removing letters until either no more letters can be removed or the new final word satisfies the length constraints. Since the order of letter is no more a problem because of the representation, the only constraint that is trying to be satisfied is the length of the final word which is calculated in equation~\ref{eq:constraints} where $w$ is the challenge word and $tl$ is the target length.

\begin{equation}\label{eq:constraints}
    c=%
    \begin{cases}
    1 & |w| \leq tl\\
    1 - \log(|w| - tl + 1) & |w| > tl
    \end{cases}
\end{equation}

If the chromosome satisfy the constraint function shown in equation~\ref{eq:constraints}, it is being tested for it fitness which tries to make some the recognized English words in the challenge word either obvious to spot or not. For example: ``CADOGT'' and ``CDAOGT'' are two different challenge words for ``CAT'' and ``DOG'' where the word ``DOG'' is easier to be noticed in ``CADOGT'' than in ``CDAOGT''. The fitness function first generates all the possible recognized English words in challenge word. For example: ``HATDEL'' is a mixture of ``HATE'' and ``DEL'' but these are not the only words in that challenge word as you can get ``HADE'', ``ATE'', ``TEL'', etc. The possible words set is split based on each word length using a certain value called the maximum sequence parameter to form two sets of words: long words and short words. The fitness function tries to minimize the number of words from the long words set that appears as a full words in the challenge word and maximize the number of words from the short words set that appear as a full words in the challenge word. 
% For example: if the max sequence parameter is 3 and we are mixing ``HATE'' and ``DEL'', ``HATDEL'' has a higher fitness score than ``DHATEL'' as all long words (``HATE'' and ``HADE'') in the first word are split while the second has ``HATE'' in order. Same with words from the short words set, the first mixture has more short words in sequence than the second one. 
Equation~\ref{eq:fitness} describes the fitness function where $f$ is the fitness function, $maxSeq$ is a the maximum sequence parameter, $chWord$ is the challenge word, and $words$ is a set of all possible words in the $chWord$.

\begin{equation}\label{eq:fitness}
    \begin{aligned}
        Long = \{w | w \in words \wedge |w| > maxSeq\}\\
        Short = \{w | w \in words \wedge |w| \leq maxSeq\}\\
        v =  |\{w | w \in Long \wedge w \in chWord\}|\\
        e = |\{w | w \in Short \wedge  w \in chWord\}|\\
        f = (1.1 - \frac{e}{|Short|}) * (1.1 - \frac{v}{|Long|}) / 1.21
    \end{aligned}
\end{equation}

After all the 10 challenges is being generated for each level, the system picks a fixed number of these challenges to apply the 2X letters for them. The 2X letter is picked by finding the least frequent letter in each source word then picking a random letter out of that list. Another note, we don't use the full corpus during generation. We define a fraction of the corpus to work with for each level so earlier levels can uses more frequent source words than later levels.

\subsection{Parameter Choices}\label{sec:parameters}
All the 30 levels in Elimination are being generated automatically by the system with minimum interference from us. We only control the difficulty of the generated level by controlling the system parameters for each level. We use the following five parameters to control the difficulty of the generated level, $corpusFreq$: the subset of the corpus used in generation (it has minimum word frequency and maximum word frequency), $sourceWords$: a list of the length of the source words being selected (usually two or three source words), $targetLength$: the length of the challenge word after mixture that is being used to calculate the constraints, $maxSeq$: the split parameter used in the fitness function to calculate the fitness, and $num2X$: the number of challenges in the level that has 2X.

The level generator of Elimination was designed to guide the user through ups and downs in difficulty of play. The idea was to alternate between challenging the user and rewarding them with successes. We selected the parameter values for each level such that the difficulty of the level increase every 5 levels then drops at the 6th level to form a saw-toothed difficulty curve~\cite{holleman2018mario,brown2010diff,starchan2018diff} which is discussed in details in section~\ref{sec:diff}. The reason for the difficulty drop is to allow players to have some breathing room after hard sequence of levels, also the combination with the increasing difficulty is to make sure that the generated levels will help the user to be in state of flow~\cite{nakamura2014concept}.
% http://thegamedesignforum.com/features/RD_SMW_2.html
% http://thegamedesignforum.com/features/RD_D2_4.html
% https://www.gamasutra.com/blogs/JonBrown/20100922/88111/Difficulty_Curves_Start_At_Their_Peak.php?print=1
% http://www.davetech.co.uk/difficultycurves
% https://www.reddit.com/r/gamedev/comments/7869o1/i_took_ben_tristems_unity_course_so_you_dont_have/
% https://csanyk.com/2016/08/appreciating-megamania/
% https://gametechdms.files.wordpress.com/2014/08/kevinlenyo_difficulty.pdf

\begin{figure}
\centering
    \begin{subfigure}[t]{0.45\linewidth}
        \centering
        \includegraphics[trim={0 0 0 22pt},clip,width=\linewidth]{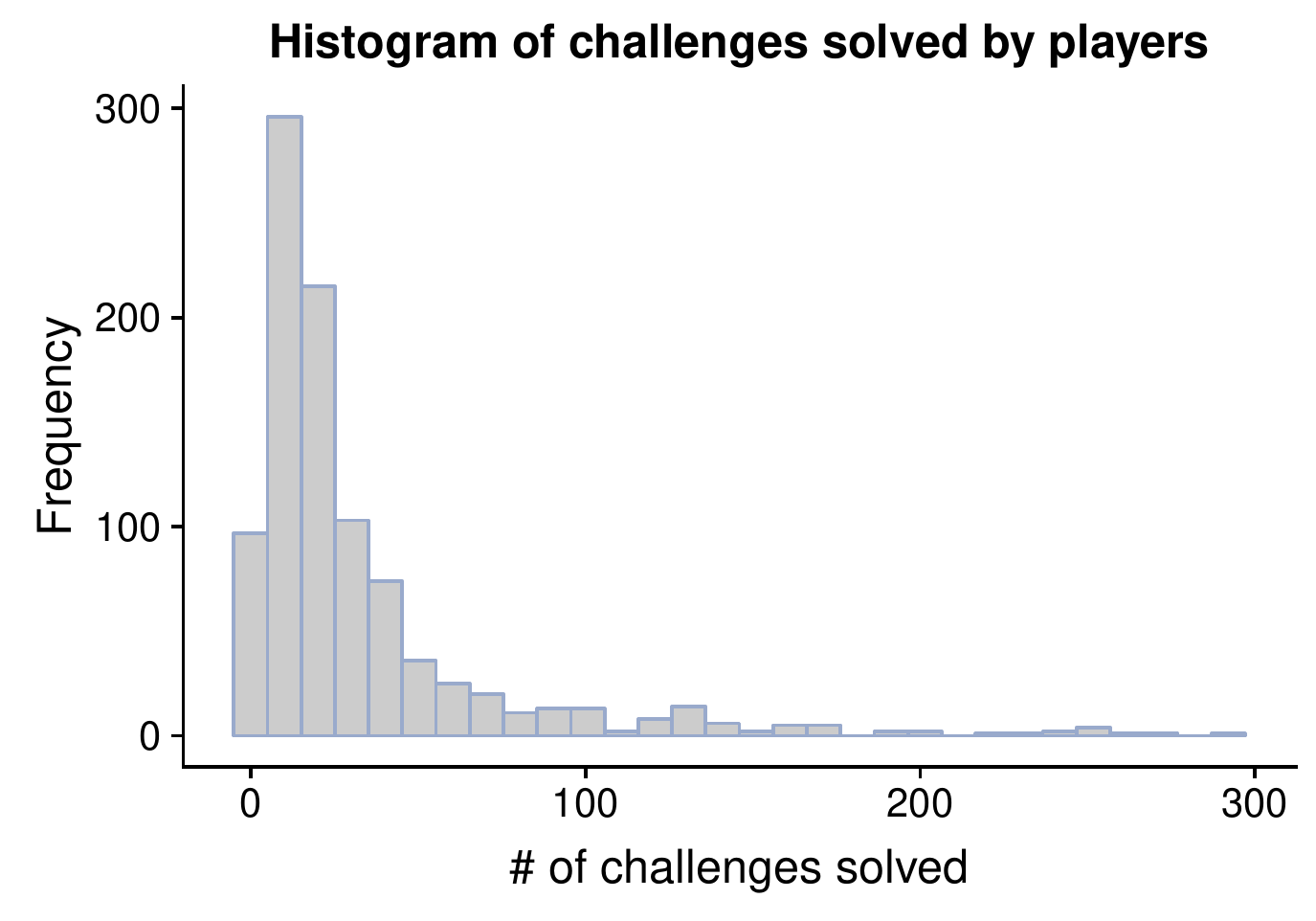}
        \caption{Histogram of challenges solved.}
        %The image is truncated at 300 challenges, the most possible without repeating a level, however some users played levels multiple times solving up to 2000 challenges.}
        \label{fig:n_challenges_per_user_hist}
    \end{subfigure}
    \begin{subfigure}[t]{0.45\linewidth}
        \centering
        \includegraphics[trim={0 0 0 22pt},clip,width=\linewidth]{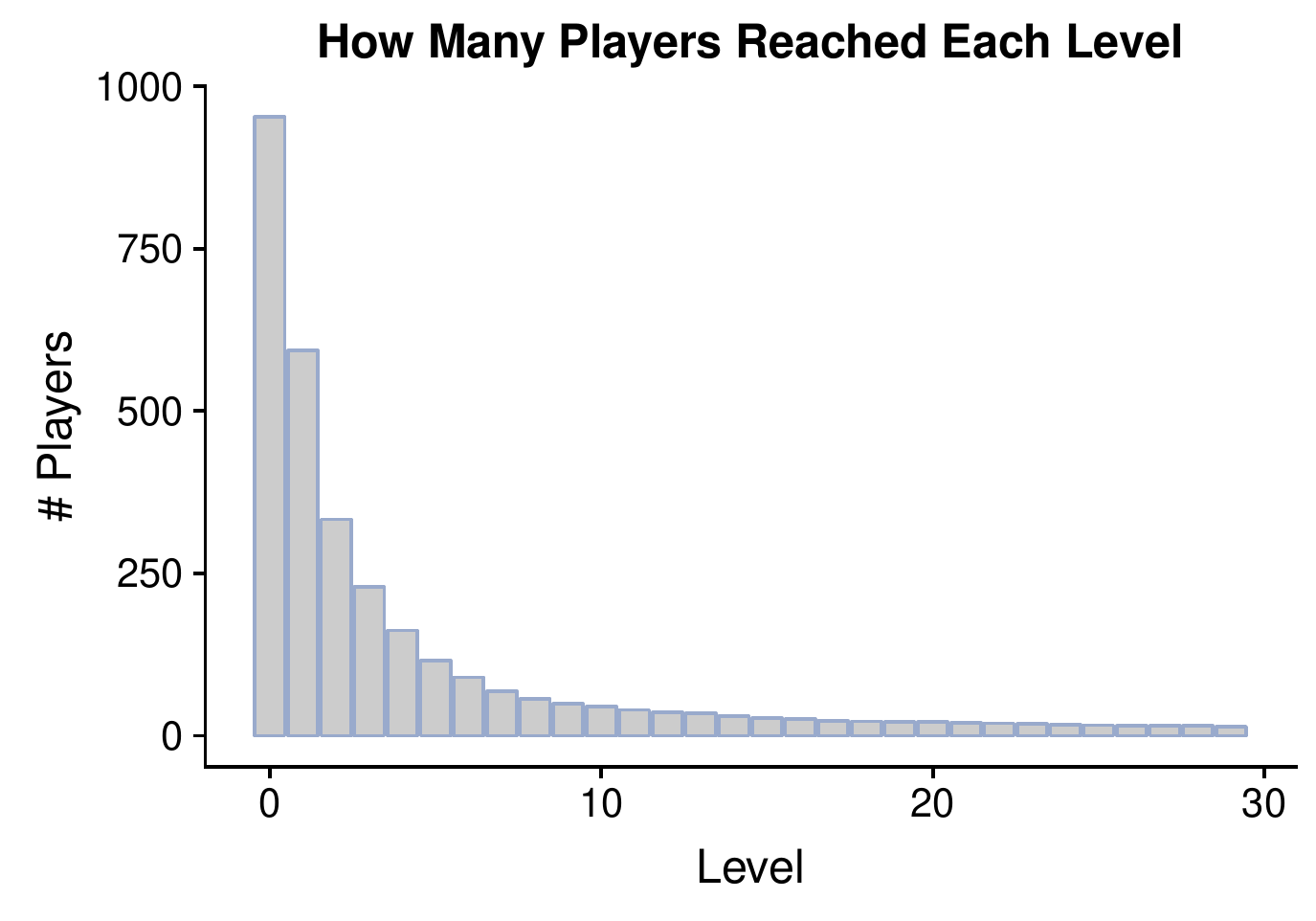}
        \caption{The number of players that reached each level.}
        \label{fig:player_dist_level}
    \end{subfigure}
    \caption{General gameplay statistics.}
    \label{fig:levels_challenges_hist}
\end{figure}

\section{Player Analysis}

Without a large sample of play-testers, the design goals of the game, and the accuracy of the level generation parameters was not fully tested before the game was released. After the public launch sufficient player data was gathered to measure the relationship between the intended design of the game and the actual reception by players.

\subsection{General Gameplay Statistics}

In total 975 users played Elimination over the course of 98 days. The dedication of these players varied from casual, people who stumbled across the game and quickly realized it wasn't for them, to the hardcore, playing levels over and over again trying to achieve perfect results. The median player solved 20 challenges, equivalent to solving all the challenge of the first 2 levels, and the most diligent player solved exactly 2000 challenges, repeating several levels over dozens of times perfecting them. Figure~\ref{fig:levels_challenges_hist} illustrates the distribution of player dedication through the number of completed challenges and the number of players that played each level.

\subsection{Validation of Difficulty Measure}\label{sec:diff}

\begin{figure}
    \centering
    \begin{subfigure}[t]{0.45\linewidth}
        \centering
        \includegraphics[trim={0 0 0 92pt},clip,width=\linewidth]{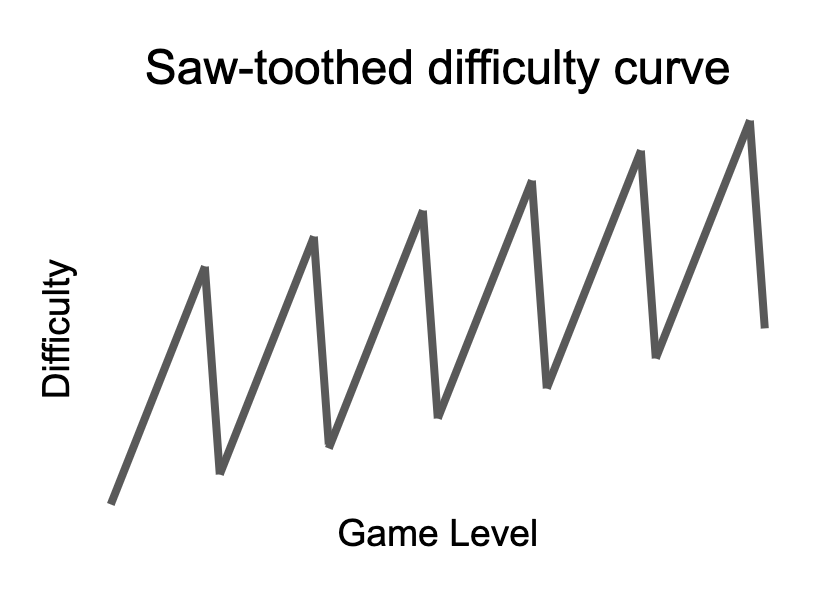}
        \caption{}
        \label{fig:ideal_difficulty_curve}
    \end{subfigure}
    \begin{subfigure}[t]{0.45\linewidth}
        \centering
        \includegraphics[trim={0 0 0 22pt},clip,width=\linewidth]{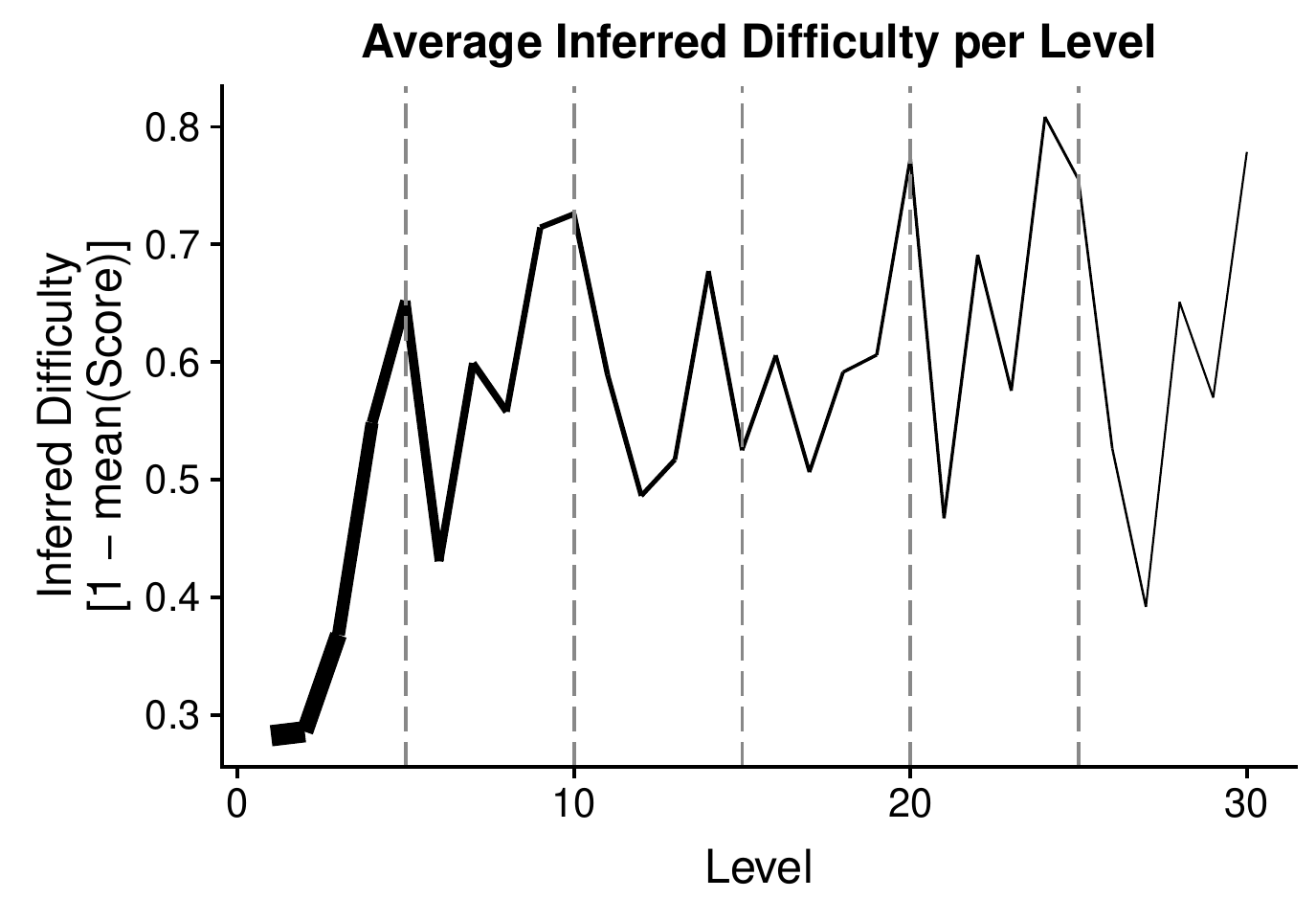}
        \caption{}
        \label{fig:observed_difficulty_curve}
    \end{subfigure}
    % \caption{Comparison between the ideal difficulty curve (Gradually increasing oscillation) on the left, and the observed difficulty curve (inferred from the mean inverse score. Quantity of observations illustrated as line thickness) on the right.}
    \caption{Comparison between the ideal difficulty curve on the left, and the observed difficulty curve on the right.}
    \label{fig:diff_curves}
\end{figure}

The shape of the difficulty curve was intended to be a saw-tooth graph (as discussed in section~\ref{sec:parameters}) similar to the one displayed in Figure~\ref{fig:ideal_difficulty_curve}. In practice it's impossible to measure the absolute difficulty of a game level, but we can estimate the relative difficulty by comparing players' score from one level to another. We can infer that levels in which players score higher are easier, and levels in which players score lower are harder. Under this interpretation we had hoped to see that player scores would follow the inverse saw-toothed pattern the game's difficulty was designed to produce. The observed outcome is displayed in Figure (\ref{fig:observed_difficulty_curve}).

To validate the efficacy of our level-generation parameters on the resultant game, we employ a linear regression to measure the predictive power of the generation parameters on the observed player scores. 
% In the best case, if our design intuition and implementation were excellent, we would see that a very high percentage of the variance of player scores are predicted by the parameters of the level generation. Other sources of variation in player scores may be due to externalities such as player ability, real-world distractions, etc. 
Our linear regression over the generation parameters took the form of $ normalizedScore \sim min(corpusFreq) + maxSeq + targetLength + num2X + min(sourceWord) $, predicting the mean normalized score of each level based on the parameters described in section~\ref{sec:parameters} where $min(corpusFreq)$ is the popularity of constituent words in the English language and $min(sourceWord)$ is the size of the smallest constituent word. Together these features explain almost 20\% of the variance ($R^2 = 19.43$) of player scores. What this model indicates is that, of the generation parameters used, the ones correlated with user score were the ones controlling word popularity, and the length of the challenge word. 

\subsection{Understanding Word Choices}

Since a level in Elimination consists of 10 independent challenges, we can get a more detailed understanding of the players' responses by focusing on the outcomes of each individual challenge. Similar to our analysis of level generation parameters, we can use a linear regression model to explain a players' selections on a word-by-word basis. 
% In this case, we are no longer comparing players' performance against hand-picked parameters, but instead against the output of the PCG. 
Rather than evaluating the efficacy of our AI algorithm, this analysis is more suited to teaching us about the way players think, while searching for words in a challenge. We analyzed the following regression $selectionRate \sim wordLength + maxSequence + has2X + splitDistance + firstOccurrence + dirtyWord$ which measures the correlation between the frequency a particular word is selected in a challenge ($selectionRate$) and the following features: the length of the selected word ($wordLength$), the largest number of letters from the selected word appearing consecutively in the original challenge ($maxSequence$), whether the selected word has a 2x bonus on one of its letters ($has2X$), the total number of letter from other words that are interspersed with letters of the selected word ($splitDistance$), the position in the challenge word of the first letter of the selected word ($firstOccurrence$), and whether or not the selected word is profane ($dirtyWord$). Together these features account for  21\% of the variance in word selection rates ($R^2 = 21.0$).

The biggest contributing features are $wordLength$ and $maxSequence$. It is fairly intuitive that words with more sequential letters directly in the challenge word are more likely to be selected, as they can be read without any visual interpretation. However, it is less obvious why longer words are more likely to be selected. There are several reasons why players might tend to select longer words. Firstly, longer words are more highly rewarded by the scoring system, selecting longer words rewards the player with a bigger score. Counter-intuitively, though, longer words may also be easier to select than shorter words. For example, longer words necessarily require fewer eliminations (clicks) to be selected than shorter words. Some elimination paths (series of clicks required to select a short word) will accidentally select larger words in the process, immediately solving the challenge. In fact, in the (somewhat rare) context of specific challenge words, there are some desired words that exist, but cannot be selected. For example: the challenge word ``AFART'', if the player tries to select the word ``FAR'' they will quickly find out it is impossible. To transform ``AFART'' into ``FAR'' requires removing the first letter `A' and the last letter `T', however neither can be removed without inadvertently creating another word (``FART'' or ``AFAR'', respectively). This property was completely unknown to us before finding the prevalence of longer word selections.

\section{Conclusion}
This paper described the design of Elimination, a word puzzle game where all the levels were generated using the FI-2Pop algorithm. The generator was controlled using five different parameters that were adjusted to reflect a saw-toothed difficulty curve. We analyzed the collected user data to validate our difficulty curve by using a linear regression model to predict the players' scores based on the generation parameters. The model was able only to explain 20\% of the data with the highest correlation for the frequency of the source words and the challenge word length. We further analyzed the data to understand more about the player word choices. We found that the length of the selected word and number of consecutive letters from the selected word in the challenge word have the most effect on the player's choice.
% In the last anaylsis, we tried to see any relation between player performance and their performance or quit rate but the data didn't have any relation between these parameters at all.

There is much room for improvements to the game, for example we can use automated tuning to adjust the parameters to fit the saw-toothed difficulty curve similar to what was done by Isaksen et al~\cite{isaksen2015exploring}. Another direction can be achieved by using Constrained Map-Elites~\cite{khalifa2018talakat} for offline generation instead of FI-2Pop to generate challenges that explores the full space of the game behaviors. This work also demonstrates the importance for player analytics of collecting detailed playthroughs containing timestamps and more detailed information rather than aggregated information.

% - Aaron Work~\cite{isaksen2015exploring}
% - Log for each challenge
% - Record timestamps

\section*{Acknowledgment}
Ahmed Khalifa acknowledges the financial support from NSF grant (Award number 1717324 - ``RI: Small: General Intelligence through Algorithm Invention and Selection.'').

\bibliographystyle{IEEEtran}
\bibliography{references}

\end{document}